\documentclass[conference]{IEEEtran}
\IEEEoverridecommandlockouts
\usepackage{cite}
\usepackage{amsmath,amssymb,amsfonts}
\usepackage{algorithmic}
\usepackage{graphicx}
\usepackage{textcomp}
\usepackage{xcolor}
\usepackage{listings}
\usepackage{url}
\usepackage{verbatim}
\usepackage{makecell}
\usepackage{booktabs}

\lstdefinelanguage{json}{
    basicstyle=\ttfamily\footnotesize,
    numbers=left,
    numberstyle=\tiny,
    stepnumber=1,
    numbersep=5pt,
    showstringspaces=false,
    breaklines=true,
    frame=single,
    backgroundcolor=\color{gray!10},
    keywordstyle=\color{blue},
    stringstyle=\color{red},
    morestring=[b]",
    morekeywords={true,false,null}
}

\def\BibTeX{{\rm B\kern-.05em{\sc i\kern-.025em b}\kern-.08em
    T\kern-.1667em\lower.7ex\hbox{E}\kern-.125emX}}

\begin{document}

\title{{\sc Story2Game}: Generating (Almost) Everything in an Interactive Fiction Game}

\author{\IEEEauthorblockN{Eric Zhou, Shreyas Basavatia, Moontashir Siam, Zexin Chen, Mark O. Riedl}
\IEEEauthorblockA{\textit{School of Interactive Computing, College of Computing} \\
\textit{Georgia Institute of Technology}\\
\{ezhou66, sbasavatia3, msiam3, zchen849, riedl\}@gatech.edu}
}

\newcommand\sysname{{\sc Story2Game}}

\maketitle

\begin{abstract}
We introduce \sysname, a novel approach 
to using Large Language Models to generate text-based interactive fiction games that starts by generating a story, populates the world, and builds the code for actions in a game engine that enables the story to play out interactively.
Whereas a given set of hard-coded actions can artificially constrain story generation, the ability to generate actions means the story generation process can be more open-ended but still allow for experiences that are grounded in a game state.
The key to successful action generation is to use LLM-generated preconditions and effects of actions in the stories as guides for what aspects of the game state must be tracked and changed by the game engine when a player performs an action. 
We also introduce a technique for dynamically generating new actions to accommodate the player's desire to perform actions that they think of that are not part of the story.
Dynamic action generation may require on-the-fly updates to the game engine's state representation and revision of previously generated actions.
We evaluate the success rate of action code generation with respect to whether a player can interactively play through the entire generated story. 
\end{abstract}


\section{Introduction}
Narratives are vehicles to communicate, entertain, and educate. Their specific purpose is accomplished as an audience experiences the narrative. {\em Interactive narratives} \cite{InteractiveNarrative} are a form of storytelling that allow players to make decisions and interact with the story, directly affecting the direction and outcome of the plot. Interactive fiction games---notably {\em text-adventure games}---are a method for players to deeply engage with these narratives by allowing players to creatively interact with a virtual world through the modality of natural language. These games provide textual descriptions of the world environment and player interaction is induced by typing text commands that their characters should take. Popular games such as {\em Zork} have rich narratives where players can affect the environment and thus move through the story in a first-person capacity. 

Interactive fictions have historically been hand-crafted, sometimes using game engines or specialized programming languages such as Inform7.\footnote{\url{https://ganelson.github.io/inform-website/}}
However, one of the emergent abilities of large language models (LLMs) pre-trained on internet content has been the ability to emulate an interactive fiction game when prompted.
For example, {\em AI Dungeon},\footnote{\url{https://aidungeon.com/}} built on top of GPT-3 \cite{GPT3}, is an infinitely generative text-adventure using AI to respond to player input.
While it showcases the ability for unconstrained interactive narratives in the interactive fiction format, it also demonstrates one of the pitfalls of LLM generated interactive narrative content: 
players are relatively under-constrained and can perform virtually any action, even those that are seemingly impossible according to the logic of the game world (e.g., being able to fly without prior indication of being able to do so), or to break the narrative flow (e.g., killing the antagonist immediately).
There is effectively no story, and thus no constraints on the player.

Many AI systems that generate interactive narratives~\cite{DynamicExperienceManagement,DynamicGenDilemmaNarratives} and interactive fictions \cite{RajStoriesAlive} 
assume an existing game engine with existing sets of pre-defined possible user actions (e.g., take, move, shoot, etc.).
This is generally assumed necessary because some underlying game engine code is required to execute in response to user actions to manipulate the ground-truth state of the world.
As a consequence, only stories that make use of the hard-coded set of actions can be generated.
This is problematic if we wish to use an LLM to generate a story that can then be executed in a game engine because the LLM may seek to introduce actions and events that are un-executable in the game engine.
The player is similarly constrained to the pre-defined set of actions.
In short: LLM-based story generation may not be {\em grounded} in the specifics of the game engine and what it is capable of representing and expressing at the code level.

In this paper, we consider {\em de novo} interactive fiction generation in which 
(a)~an LLM-based story generator produces a story without constraints, and 
(b)~an interactive game engine is then dynamically constructed specifically for the purpose of playing out the story, after which 
(c)~the player can then creatively interact in the text game, with novel actions dynamically generating events and consequences that are grounded within the world's logic.
We assume a {\em minimalistic} text game engine that implements basic data structures for maintaining the state of the world, including world locations (rooms), objects, NPCs and a player agent. 
The world is populated according to the needs of the story and the actions that the game engine can execute are also generated according to the story. 
Later, if the player wants to perform actions that are not part of the story, these actions are generated at the time that the user attempts to execute them, and are implemented in a way that remains consistent with the generated world and existing objects, locations, and actions. 

We introduce the \sysname{} system, which implements a multi-staged generation process.
First, an LLM-based story generator produces a story.
We use a story generation technique 
in which an LLM iteratively constructs a story comprised of actions with preconditions and effects.
The preconditions and effects ensure causal logical story progression. 
We additionally use the preconditions and effects to determine the semantics of how each action in the story can be executed.
For example, the preconditions of \texttt{unlock the chest} would indicate that the player would need to have a key at the onset of the action, and  at the end of the action's execution the chest would be opened.
The second stage generates the locations in the world and characters and objects that are relevant to the story. Our implementation of world generation is relatively simplistic and not an emphasis of the work presented in this paper.
The third stage is to generate code for the actions in the story that will execute when the player types in an action.
The code must update the basic data structures of the player, location, and other objects that are consistent with the semantic intent of the action. 
For example, a \texttt{buy ticket} action should only be executable in the relevant location, if the data structure representing the player has money listed in their inventory slot and the player inventory data structure is updated to have a ticket but no money. 
The actions must also enable the player to perform subsequent actions according to the generated story in the order they appear in the story.

A player that does not know the story will want to execute actions that are not part of the story and thus is not anticipatable by the above generation process.
We also introduce a {\em dynamic action generation} process whereby we generate new actions, which may require the creation of new objects (e.g. a torch to burn something) or the creation of new character attributes (e.g., a strength attribute to lift something heavy).
Existing actions may need to be revised to take into consideration the existence of {\em a priori} unrealized objects and attributes.

We evaluate the initialization of the text world and dynamic action generation, specifically focusing on successful compilation and a logical implementation of novel events that matches a human's commonsense understanding within the grounding of the world. Using a variety of short-stories of differing complexity and setting, we create a multitude of novel actions for existing objects in the game and measure the efficacy and coherence of the dynamic action generation.

\section{Related Work}

\textit{Interactive narratives} \cite{InteractiveNarrative}
are those in which a story plays out but the player can take actions to change the direction or outcome of the story. 
Early
\textit{AI generated interactive narratives} use an AI story generation process to create stories that play out in a pre-existing game engine.
Early interactive narrative generators used planning \cite{riedl:saretto:aamas2003} \cite{CavazzaNarrativePlanning} \cite{FacadeSystem}, search \cite{NelsonNarrativeSearch} \cite{WeyhrauchNarrativePlanning}, or autonomous agents\cite{autoagents}. 
\cite{Scheherazade} and \cite{SwansonMLforIF} used machine learning techniques to generate choose-your-own-adventure interactive narratives for example text corpora.

The Game Forge system~\cite{Hartsook2011TowardSupportingStories} is similar to our work in scope.
It uses a symbolic story generator to generate a story as a sequence of actions, a genetic algorithm to evolve a map, and performs some basic configuration of action templates.
\sysname{} uses an LLM for all generation stages and focuses specifically on story-based action generation for the story and to support dynamic player actions. 

With the onset of Large Language Models, \cite{RajStoriesAlive} used LLMs to analyze existing stories and novels and construct interactive fiction worlds with locations, characters, and objects from those stories. 
This system did not generate the stories and it only used existing text world actions such as \texttt{move} and \texttt{look}.
The Word2World system \cite{Nasir2024Word2WorldGS} uses an LLM to generate a story and then create a 2D world populated by characters and objects.
This system relied on a pre-specified set of sprites and used consistency checking algorithms to make sure sprites a placed correctly.
The Starling system \cite{Basavatia2024STARLINGST} similarly generated stories by prompting an LLM simple story ideas such as "cook pasta" that represent everyday actions. These stories were converted into Inform7 code and compiled into playable IF games used for pre-training RL agents.

Our work is consistent with \cite{LLMsAndGamesSurvey}, a set of desiderata on the roles that LLMs can take in games, including a {\em Game Master} that creates the plot of the story and prepares it to be played, and {\em Automated Designer} where the LLM modifies and adapt the game by mapping natural language outputs to game-specific formats.

\section{\sysname}

\begin{figure*}[t]
    \centering
    \includegraphics[width=\textwidth]{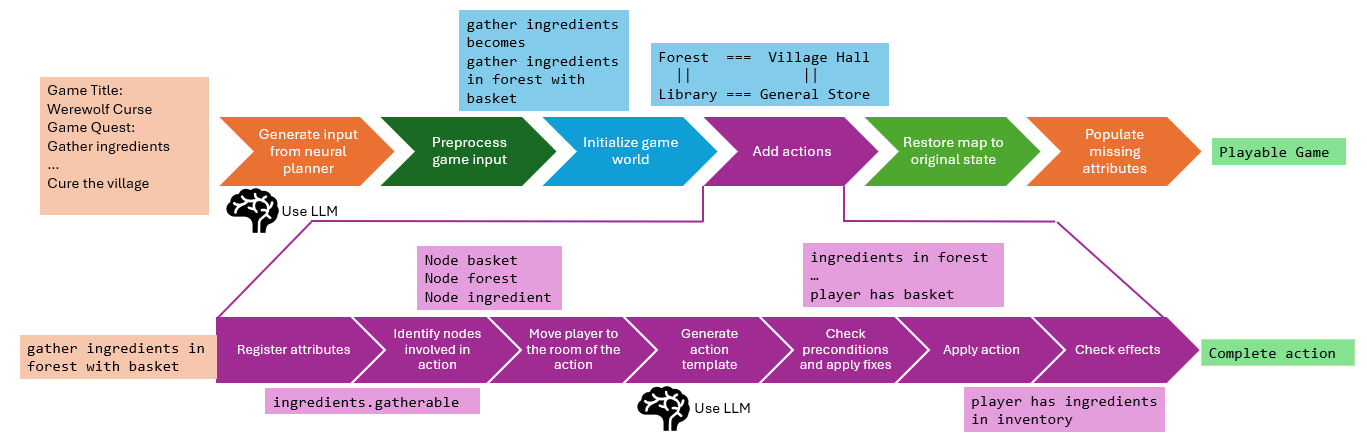}
    \caption{Game Generation Pipeline.}
    \label{fig:V1Diagram}
\end{figure*}

This section discusses the framework for our game-building agent, \sysname. 
For an overview of the game initialization pipeline, refer to Fig. \ref{fig:V1Diagram}.

Our \sysname{} generates content for a minimalistic text game engine.
Consistent with game engines/languages such as Inform7, TextWorld \cite{TextWorld}, LIGHT \cite{Urbanek2019LearningTS}, and others, a minimal amount of data about world state must be maintained.
Our default game engine maintains data structures for the world locations and map layout, objects, and the player. 
The player data structure, in particular, has an inventory list.
The engine also provides a basic currency system, which is common in interactive fictions.
We provide three default actions: \texttt{look}, \texttt{open inventory}, and \texttt{go to [room]}, that are universal. These core features serve as scaffolding for the text-based game.

\subsection{Generating Stories with Preconditions and Effects}
\label{sec:story-gen}

The first stage of game generation is to generate a story. 
An LLM is prompted with a title for the story, a sequence of 2-6 main events, as well as a description of the the goal and the setting of the world. 
From this, the LLM---GPT-4o-mini---will generate between 5-18 events. 
Table \ref{tab:story_example} shows an example input to the story generation process input.
The resulting sequence of sentences will be, due to the nature of interactive fictions, descriptions of actions that the player will perform. 

\begin{table}[b]
    \centering
    \renewcommand{\arraystretch}{1.2} 
    \caption{Story Generation Example}
    \begin{tabular}{|c|p{5.5cm}|}
        \hline
        \textbf{Title} & Guardian's Heirloom \\ 
        \hline
        \textbf{Main Quest Line} & 
        ``Infiltrate the rival clan's fortress to retrieve the heirloom.'',
        ``Solve puzzles and traps within the fortress to reach the treasure.'',
        ``Face off against the rival clan's skilled warriors in combat.'',
        ``Return the guardian's heirloom to its rightful resting place.'' \\ 
        \hline
        \textbf{Description} & The player is the last descendant of an ancient guardian lineage and must retrieve a powerful family heirloom stolen by a rival clan. \\ 
        \hline
    \end{tabular}
    \label{tab:story_example}
\end{table}

A sequence of sentences describing character actions, however, leaves a lot of semantics of what happens in an interactive environment implicit. 
Following Yu et al.~\cite{Ye2022NeuralStoryPlanning}, 
we prompt the LLM to identify all necessary preconditions to perform an action, as well as any in-game effects this may have on the world. 
{\em Preconditions} are statements about the state of the world that must be true for an action to be executable.
{\em Effects} are statements about how the state of the world will be different if an action is executed.
While both the planner by Yu et al. and our story generator create preconditions and effects for each action in a story, our approach differs by generating a text story first and annotating preconditions and effects after. 
Our story generator is simple and not the main contribution of this work; more sophisticated story generators could be used instead as long as they produced preconditions and effects.

Preconditions are split into three groups.
{\em Fundamental preconditions} involve generic location and inventory checks.
{\em Additional preconditions} encapsulate a variety of possible 
requirements, most commonly represented through the application and comparison of custom attributes (i.e. a chest being unopened for it to be openable, or for the player to have a certain “deception” attribute to distract a guard”).  For example, given a simple sentence such as ``The adventurer burns the notebook with a match at the living room'', fundamental preconditions would involve: ``adventurer in living room'', 
``adventurer has match'', 
``adventurer has notebook'', while additional preconditions would just include: \texttt{notebook.burned == False}. 
{\em Preceding events} are requirements that a previous action must be completed before the current one can be executed. 
They require a check of all previous sentences to see if they are required for the current action. Otherwise, action effects are not necessarily guaranteed to chain with future actions' preconditions.

The LLM also determines the type of effect that the event/action has on the world. 
Effects are divided into four  categories: (1)~{\em movement} from one location to another, (2)~{\em setting} of an attribute, (3)~{\em creation} of a new object, or (4)~{\em removal} of an existing object.
As with fundamental preconditions, these categories of effects are nearly universal across the interactive fiction game genre.

\subsection{World Generation}
\label{sec:world-gen}

The game world is represented as a graph where nodes  represent any object within the story.
All nodes are categorized as \textit{Player}, \textit{Character}, \textit{Item}, \textit{Room}, or \textit{Container} classes at the code (python) level. 
The Player and Character classes are similar except the Player class has an addition \texttt{observation()} method that gets called when the agent executes the \texttt{look} action.
Players and Containers have inventory slots.
Room objects can contain objects of all classes, including other rooms (e.g. a forest has a cabin in it, where forest and cabin are distinct visitable locations). All nodes can have attributes applied to them, except for rooms (such as exploding a room, thereby making it inaccessible).

Rooms are identified in the story and new Room objects are instantiated on a grid. Starting with the first room, each successive room in the story is placed randomly to the north, south, east, or west of the previous room.
All actions in the story are associated with a room.
If the action does not directly reference a room, we assume it is associated with the most recently reference room.
Thus, all Items and Characters participating in an action are also associated with, and placed within a Room when they are instantiated as objects in the graph.
There is no reference to where the item may be within the room respective to other objects; all are simply inside  it.

\subsection{Game Engine Generation}
\label{sec:engine-gen}

The translation of actions from the story (Section~\ref{sec:story-gen}) into code that implements the actions is performed by an LLM. 
Each action is built within the game engine in the order that it appears in the story. 
Preconditions are directly translated into code that checks for relevant information in the game engine. 
As mentioned, each action is associated with fundamental preconditions, additional preconditions, and effects. Outside of preceding events, all other fundamental and additional preconditions are categorized into one of the following: 
(1)~\textit{node location check} - that the required nodes are in the correct room; 
(2)~\textit{inventory check} - that the player has said item/s; or 
(3)~\textit{node attribute check} - that the attribute of a node is satisfied.
Each effect is directly translated into an in-game operation that changes the data structures within the game engine.
These operations correspond to the four effect categories mentioned previously, altering the game state by: moving a node's location, setting an attribute of a node to some value, adding a node into a selected room, and deleting a node from the world state.
Table~\ref{tab:action-example} shows an example of an action after translation from story.

\begin{table}[b]
    \caption{Action Example}
    \label{tab:action-example}
    \centering
    \renewcommand{\arraystretch}{1}
    \begin{tabular}{| p{0.3\linewidth} | p{0.6\linewidth} |}
        \hline
        \textbf{Action} & player, distract the \{guard\} at \{dungeon\} \\ 
        \hline
        \textbf{Preconditions} & \{\{player at dungeon\}, \{guard at dungeon\}\} \\ 
        \hline
        \textbf{Effects} & Set \{guard\}.distracted to True \\ 
        \hline
        \textbf{Preceding Events} & “Set \{bush\} on fire at \{forest\}” \\ 
        \hline
        \textbf{Display} & You distracted the guard. \\ 
        \hline
    \end{tabular}
\end{table}

\begin{figure*}[t]
    \centering
    \includegraphics[width=\textwidth]{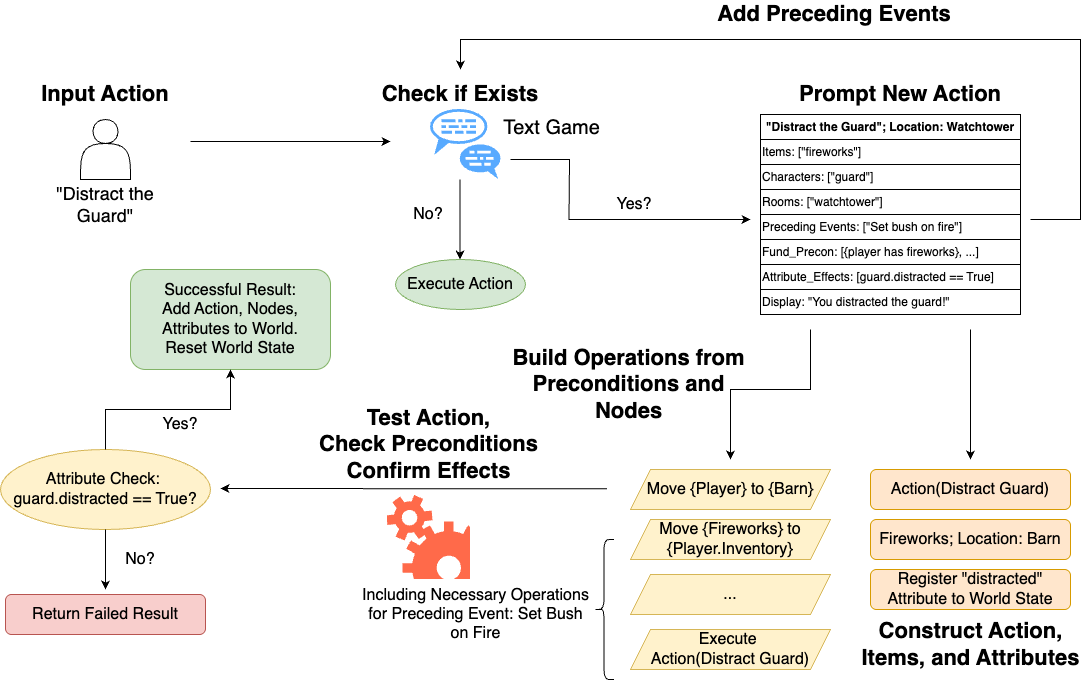}
    \caption{Dynamic Action Generation Pipeline.}
    \label{fig:dynamic-actions}
\end{figure*}

\section{Dynamic Player Action Generation}

The game engine generated in the previous section enables the player to execute every action in the story in the prescribed order.
But the player does not know the story and may seek alternative ways to make progress and explore the world. 
The player may attempt actions that are not part of the story and thus do not exist in the game engine. 
Instead of automatically informing the player that they cannot do what they want, we devise a means to allow users to perform new actions that weren't created when the game was initialized. 

When the player attempts an action that does not exist in the game engine, our engine will create the new action on the fly and incorporate it into the game engine.
There are several critical challenges to this.
(1)~The desired action may require unforeseen preconditions that reference objects in the world that don't exist yet.
For example, \texttt{break bucket} may require a hammer to reasonably execute, but there may not be a hammer already existing in the world.
(2)~The desired action may invoke attributes on objects that were not previously considered and thus not part of the game engine.
For example, \texttt{break bucket} may want to set a \texttt{broken = True} tag on the player's bucket object, but that attribute does not exist in the context of the world.
(3)~The new attributes may be relevant to other, existing actions.
For example, a broken bucket may not hold water, so a future existing event such as \texttt{fill bucket with water} may need to be updated to check for the \texttt{broken} tag as a precondition.

Our dynamic player action generation process is similar to the action code generation in Section~\ref{sec:engine-gen}, however we do not have preconditions and effects from the story to guide the process.
The LLM is now open to infer preconditions and effects pertaining to objects and attributes that do not exist in the world. 
This opens up the ability for the user to generate side storylines, or find interesting and unique ways to reach the goal.
The player will not necessarily be able to jump directly to the conclusion of the story because actions that might surpass parts of the story will have preconditions that may require the player to do other actions (e.g., find a torch).
These actions may in turn require other actions, and the player will effectively be ``off script'' and forging their own emergent story (cf. \cite{Kriegel2008EmergentNA}).

We note that there is no guarantee that new actions will not break the completeness of the game, making it impossible to finish the story.
Riedl et al.~\cite{riedl:saretto:aamas2003} describe techniques for ensuring that player actions cannot interfere with story progress ({\em intervention}) or for incorporating new actions into a storyline ({\em accommodation}). 
These techniques analyze the effects of non-story actions on story progression and are thus compatible with our game engine.

The dynamic action generation process is visualized in Fig.~\ref{fig:dynamic-actions}.
We prompt the LLM for the following information: 
items needed to complete the action;
characters involved;
rooms where the action can be performed;
preceding events;
fundamental preconditions;
effects;
and a text string to display when executing.
See Appendix~\ref{appendix:prompting-dynamic} for the dynamic action generation prompt, and an example action.

We restrict new actions to only be those that operate on existing items in the game.
This ensures that the preconditions and effects have at least one object that already exists in the world against which it can be grounded. 

\subsection{Essential Object Preconditions}

Any essential objects or characters that will be needed to perform the action and do not currently exist (e.g., scissors for cutting, a torch for burning) are created by giving them a name and a location. Locations are chosen randomly. Other placement procedures are possible, such as identifying a room the player has never entered, or a room that is semantically related to the object.

\subsection{Fundamental and Additional Preconditions}

Fundamental and additional preconditions represent the necessary locations and attributes of the relevant objects in the action. The process for translating these preconditions into matching in-game representations is identical to our game engine generation (See Section \ref{sec:engine-gen}).
Preconditions may also involve new attributes that did not exist. For example, the player may require a particular strength or dexterity, or a chest must be unopened to be opened.
If these attribute slots did not exist, they are added to the associated object and given a default value. This default is determined by the LLM, as it can differ depending on the attribute. Attributes must either be binary (assigned a True or False value), or integer based, between a scale of 0 to 10. The player may now issue actions that increase the attribute to satisfy the precondition. For example, a player may want to \texttt{{Fight the guard}.} In doing so, a Strength attribute might be assigned and a player may need to train to increase their Strength to a threshold to fight the guard.

\subsection{Preceding Event Preconditions}

While not strictly necessary, we find the resulting player experience more interesting when some new actions require preceding actions. 
For example, if a player wants to drive a broken-down car, they may need to hotwire the car, add gasoline, and fix the engine. Whenever preceding events are indicated as a precondition for a new action, we check to see if the preceding events exist. If not, we recursively create these preceding events as actions using an identical process as before. We limit the depth of preceding events to one so that there is not an infinite cascading series of events that need to be completed. 

\subsection{Attribute Effects}

Effects may change an attribute of an object or of the player. Since the state of the world is now being changed in a way that was not anticipated initially, we need to consider whether the action can change the world state in a way that has consequences for other actions.
If the new action results in the creation of a new attribute for an object, we iterate through the other actions that involve said object. We then ask the LLM to determine whether this new attribute is relevant to each prior existing action. If it is, a precondition for the attribute is created and appended to the action, which may result in the new action making it impossible to execute the latter one. This is by design; by enabling the user the liberty to construct their own storylines and paths to achieve the goal, it is just as important to ensure that these new actions have legitimate impact on the original world state. Otherwise, it would serve a similar role as an entirely detached short story.

\section{Evaluations and Results}

\begin{figure}[t]
    \centering
\includegraphics[width=\columnwidth]{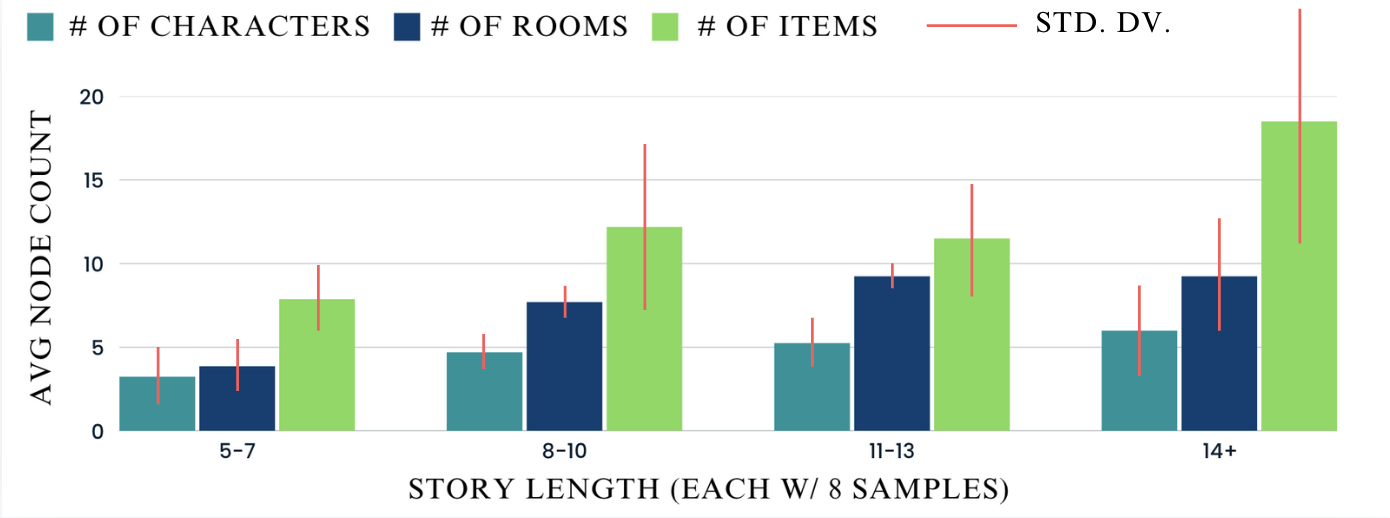}
    \caption{Average node counts of different types based on story length}
    \label{fig:game-size}
\end{figure}

We evaluate our game engine generation and dynamic player action generation separately.
For all experiments we use GPT-4o-mini as for all LLM use.
For both, we generate stories and bucket them according to their length (5-7, 8-10, 11-13, or 14+ actions).
Each length group has 8 stories.
Fig.~\ref{fig:game-size} shows statistics about how stories of different lengths produce worlds of different average sizes. 
The number of characters per story does not increase disproportionally as the length of the story grows. 
Between the shortest and longest length group of stories, whereas the number of characters increases by $89\%$ with story length, the number of rooms increases by $142\%$ in the number of rooms, and the number of items increase by $122\%$.
The fewest number of items observed was 12 and the most items in a story was 35.
Stories also grow in complexity with story length, as measured by the number of preceding action requirements per action.
Short stories (5-7 actions) typically require only a single preceding event, while longer stories may involve actions with multiple previous event requirements.

\subsection{Game World Initialization Experiments}

To evaluate our world engine generation, we test the comprehensiveness and correctness of the initial game state. 
We test each action in the game engine with respect to whether it enables the player to make progress through the story.
Starting with the first action in the story, we check whether the fundamental preconditions are satisfied by the current game state.
That is, the nodes referenced pass location and attribute checks. 
The generated story often does not not include \texttt{Move} actions;
movement to the location where an action must be performed implicitly. 
Thus, if the player or any other object is not in the required location, we force the player and/or object to move to the location that the action takes place. 
After forcing movement, if no precondition check fails, we mark the action as successful.
We then execute the action's effect operations, which change the game state graph.
We then progress to test the next action in the story until all actions have been marked as successful or unsuccessful.

\begin{table}[b]
\centering
\caption{Action generation success rates.}
\label{tab:compilation_metrics}
\resizebox{\columnwidth}{!}{%
\begin{tabular}{|c|c|c|}
\hline
\textbf{Story Length} & \textbf{\% Sentence Compilation} & \textbf{\% Fully compiled stories} \\
\hline
5-7  & 0.972 & 0.875\\
8-10 & 0.937 & 0.75\\
11-13  & 0.928 & 0.625\\
14+  & 0.942 & 0.75\\
\hline
\end{tabular}
}
\end{table}

In Table~\ref{tab:compilation_metrics} we show the percentage of actions that were flagged as successful.
We also report the number of stories for which all actions were successful. 
The vast majority of sentences successfully compile and are added to the game state.
The most common reason for sentence/action compilation failure is object misidentification by the LLM. Most commonly, this is when rooms, characters, or items are described with an adjective that may not originally have been in its description. 
For example, the LLM might at times refer to a \texttt{Key} but at other times refer to the same key as a \texttt{Metallic Key}. 
In cases such as this, the compiler doesn't know how to resolve the ambiguity as to whether these objects are the same. Certain adjectives are also sometimes essential to differentiating between similar items, so removing all adjectives leads to similar issues.
As the story length grows, the number of failures increases slightly, as would be expected because each action has a slight chance of object misidentification.
Consistent with a high per-action success rate, a majority of stories do not have any actions with compilation issues.

The compilation errors are not necessarily fatal to the player, as the player is unaware of the story actions required of them and can use dynamically generated actions to find alternative ways to make progress through the story and recover.

\subsection{Dynamic Action Generation}

For dynamic action generation, our main evaluative metric is judging how complete and coherent the new action is. For every item and character in a story, we prompt the LLM to provide three verbs that can be applied to it.
We ensure that all generated verbs do not already exist as actions in the story and re-prompt if necessary.
For example, given the object \textit{map}, we obtain the verbs, \textit{read}, \textit{fold}, and \textit{tear}. 
We then input each new verb-object combination into the game as a player action (e.g., \texttt{read map}).

\begin{figure}[b]
    \includegraphics[width=\columnwidth]{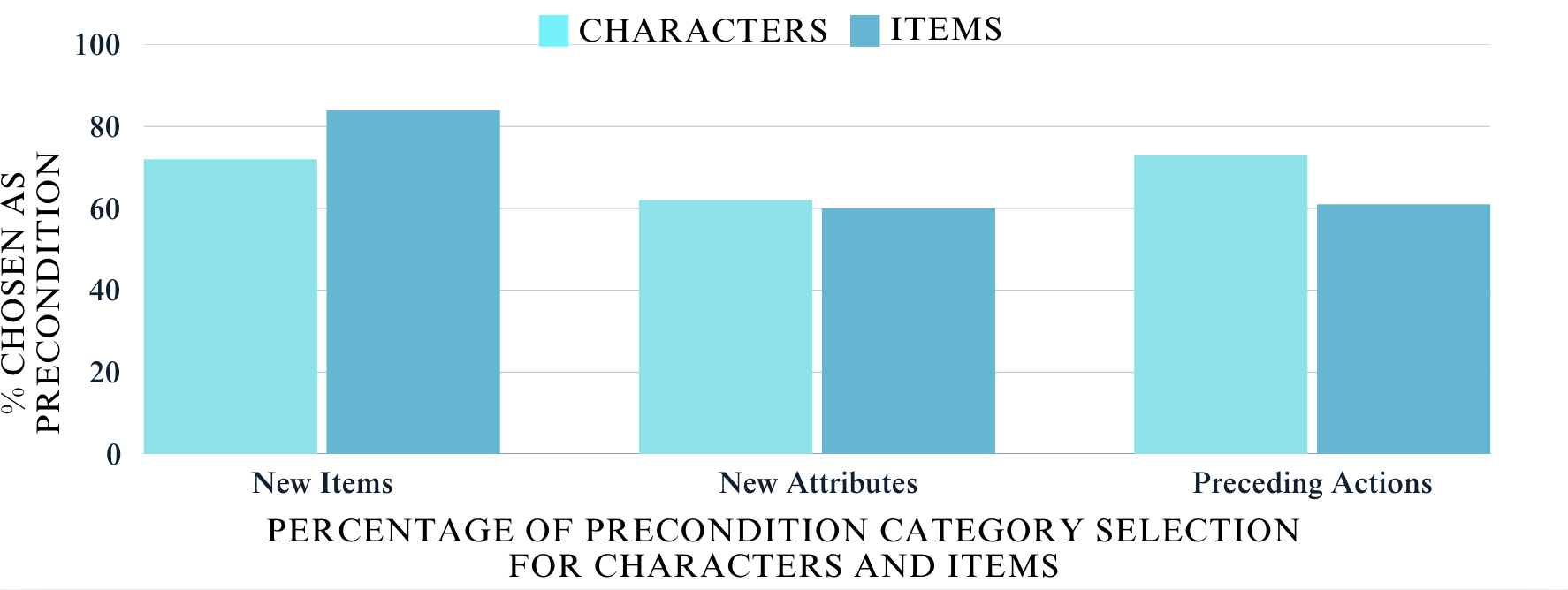}
    \caption{Percentage of times each category was chosen as a precondition. 30 total items and characters considered, each applied with 3 novel verbs.}
    \label{fig:precond-types}
    \centering
\end{figure}

By testing on 15 items and 15 characters over 5 different stories, Fig. \ref{fig:precond-types} shows the percentage at which each precondition category was considered: those that featured a new object being created, a new attribute being created, or referencing a new preceding event. It is important to note that multiple types of preconditions can be added to the same action.
We see that new items are the most common precondition applied; this is likely because many actions often involve the player using an item in a certain way to execute an action. On the other hand, the fact that characters often have new item preconditions attributed less often to them may be because verbs related to speaking or questioning do not require any items, whereas items are often interacted with physically. New attributes are the least common precondition chosen, though they are still used in over half of the cases for both characters and items. There is a larger prevalence of preceding action preconditions being chosen with characters, as novel actions applied to characters can be more vague than ones applied to objects. For example, given a "guard", one can \textit{distract} or \textit{trick} a guard. Compared to actions such as \textit{kill}, which can simply require a \textit{sword}, these actions may require previous events to have occurred alongside certain items to execute them. 

We measure the percentage of actions that successfully compile in two ways.
\textit{Compilation success} is the percentage of actions dynamically generated that are successfully built in the game and have certain preconditions that can be fulfilled by progressing through the world state.
\textit{Semantic success} is the percentage of dynamically generated actions that result in code that matches human commonsense expectations of the effects. We consider what types of preconditions are necessary to execute an action, as well as what effects the action may have on the world, and see if they match a human's understanding of the action.
For example, if we generated the action, \texttt{drop the bag}, we would expect the execution to involve the bag being removed from the player's inventory and placed into the room the player is currently located in. But if the action merely applied a \texttt{dropped} attribute to the bag while keeping it in the user's inventory, we would find that the bag would follow the player around when they moved between rooms.
We measure semantic success subjectively by visually inspecting the code from each dynamically generated action.

\begin{figure}[t]
    \centering
    \includegraphics[width=\columnwidth]{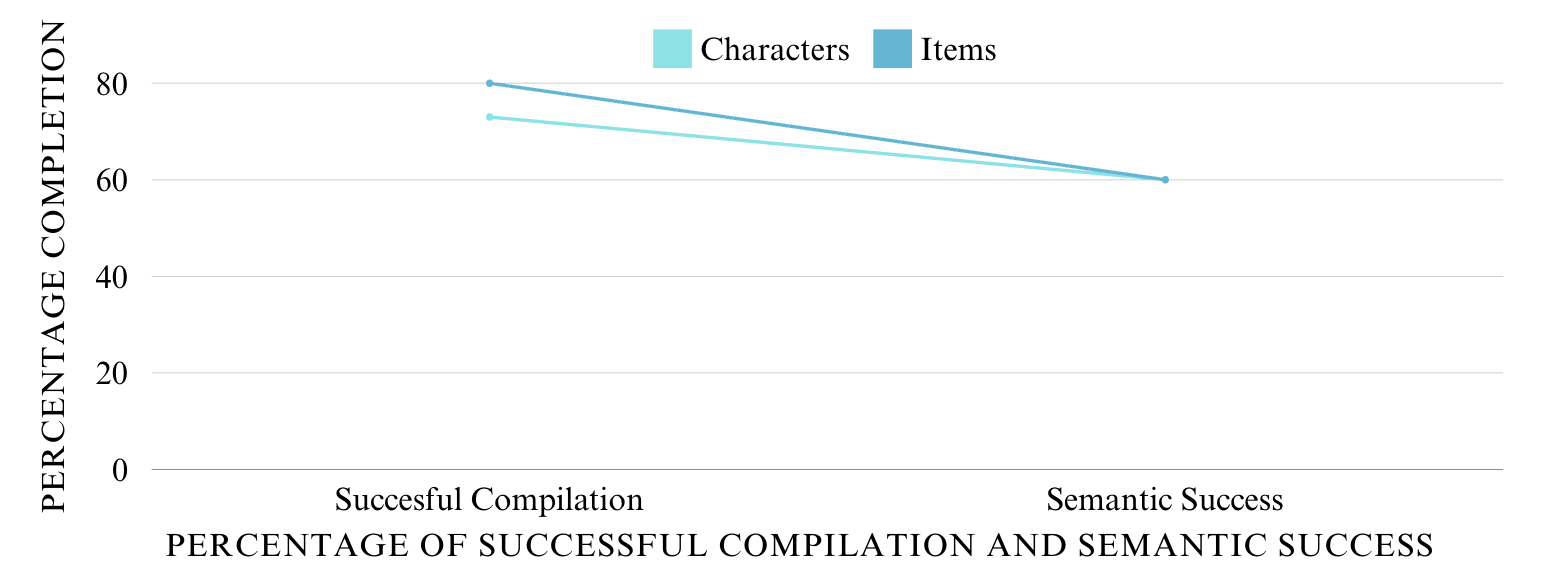}
    \caption{Novel Action Generation Confirmation Percentiles}
    \label{fig:success-percentiles}
\end{figure}

Fig.~\ref{fig:success-percentiles} shows compilation and semantic success rates on 90 actions.
We break the rates out according to whether the actions are being performed on objects or characters. 
Compilation success is a necessary precursor for semantic success, so semantic success can not be greater than compilation success.
Thus, the relative drop-off is the key measure for semantic success.
We see an $\sim 80\%$ success rate for dynamically generated actions compilation.
Semantic success decreases by $\sim 20\%$.
Dynamic action generation failures occurs most commonly due to actions that cannot be effectively represented in our game architecture. 
Most notably, while executing actions \textit{on} characters and items tends to compile effectively, executing actions \textit{with the use} of characters and items is much harder to represent. In cases where these actions do compile, the semantic success is often misguided. For example, with the phrase ``illuminate the forest with the flashlight'', a human would expect that the room ``forest'' would now have an \textit{illuminated} property, potentially with any \textit{hidden} or \textit{secret} items now visible. Rooms cannot have properties tied to them, to avoid the user making a room entirely impassable (by applying some sort of "destroyed" attribute), and as such there is no structural connection between room properties and the effects that it may have on any objects within it. Actions like these subsequently are difficult to parse and depict semantically within the game, which results in the total semantic success rate of $\sim 60\%$.

\section{Conclusions}

Generating a playable game from near-scratch is a challenging task.
Stories generated from LLMs can involve arbitrary actions and events that may not exist in a game engine.
Conversely, imposing a fixed, {\em a priori} set of actions and events on the story generator can severely limit the creativity and value of using an LLM. 
In this paper, we introduce a means of generating the code that can execute user actions for a minimalistic interactive fiction. 
Our technique leverages preconditions and effects of actions in the story to provide context for how the code for the actions should operate in the game engine. 

Players do not know the actions they are supposed to execute in order for a story to progress.
We thus provide a technique for dynamic generation of actions just-in-time when the player invents an action. 
New actions may require new objects to be created in the world, and also previously generated aspects of the game engine may need to be re-considered, such as attributes on characters and objects, which may in turn require updates to existing actions in the game engine.
This keeps new actions grounded.
Unlike under-constrained LLM-powered interactive fictions, the player may not be able to do everything they want to do right away without triggering additional preconditions that require additional player action to resolve.

The current state of \sysname{} combines deliberative storylines with emergent gameplay.
Future work may consider more complex analysis of when dynamically generated actions interfere with the ability to progress in the story, when they complement the story, and when they should be prohibited.

This paper operates within the  interactive fiction game genre because player actions are expressed through language and we can elide graphics and spatial arrangements of visual assets; the human imagination is the best graphics rendering engine.
However, as graphical generative capabilities progress, diffusion-based game engines~\cite{Valevski2024DiffusionMA} will need to consider actions that are not set in advance. 
Our work provides a step forward towards (almost) fully generated game worlds where users have large amounts of creative autonomy, while still maintaining a ground truth world state for consistent and coherent story-driven game experiences.

\bibliographystyle{IEEEtran}   
\bibliography{references,lab}

\begin{thebibliography}{10}
\providecommand{\url}[1]{#1}
\csname url@samestyle\endcsname
\providecommand{\newblock}{\relax}
\providecommand{\bibinfo}[2]{#2}
\providecommand{\BIBentrySTDinterwordspacing}{\spaceskip=0pt\relax}
\providecommand{\BIBentryALTinterwordstretchfactor}{4}
\providecommand{\BIBentryALTinterwordspacing}{\spaceskip=\fontdimen2\font plus
\BIBentryALTinterwordstretchfactor\fontdimen3\font minus
  \fontdimen4\font\relax}
\providecommand{\BIBforeignlanguage}[2]{{%
\expandafter\ifx\csname l@#1\endcsname\relax
\typeout{** WARNING: IEEEtran.bst: No hyphenation pattern has been}%
\typeout{** loaded for the language `#1'. Using the pattern for}%
\typeout{** the default language instead.}%
\else
\language=\csname l@#1\endcsname
\fi
#2}}
\providecommand{\BIBdecl}{\relax}
\BIBdecl

\bibitem{InteractiveNarrative}
M.~O. Riedl and V.~Bulitko, ``Interactive narrative: An intelligent systems
  approach,'' \emph{AI Magazine}, vol.~34, no.~1, pp. 67--77, 2013.

\bibitem{GPT3}
T.~B. Brown, B.~Mann, N.~Ryder, M.~Subbiah, J.~Kaplan, P.~Dhariwal,
  A.~Neelakantan, P.~Shyam, G.~Sastry, A.~Askell, S.~Agarwal,
  A.~Herbert{-}Voss, G.~Krueger, T.~Henighan, R.~Child, A.~Ramesh, D.~M.
  Ziegler, J.~Wu, C.~Winter, C.~Hesse, M.~Chen, E.~Sigler, M.~Litwin, S.~Gray,
  B.~Chess, J.~Clark, C.~Berner, S.~McCandlish, A.~Radford, I.~Sutskever, and
  D.~Amodei, ``Language models are few-shot learners,'' \emph{CoRR}, vol.
  abs/2005.14165, 2020.

\bibitem{DynamicExperienceManagement}
M.~Riedl, A.~Stern, D.~Dini, and J.~Alderman, ``Dynamic experience management
  in virtual worlds for entertainment, education, and training,''
  \emph{International Transactions on Systems Science and Applications -
  ITSSA}, vol.~4, 01 2008.

\bibitem{DynamicGenDilemmaNarratives}
H.~Barber and D.~Kudenko, ``Dynamic generation of dilemma-based interactive
  narratives,'' \emph{Proceedings of the AAAI Conference on Artificial
  Intelligence and Interactive Digital Entertainment}, vol.~3, no.~1, pp. 2--7,
  Sep. 2021.

\bibitem{RajStoriesAlive}
P.~Ammanabrolu, W.~Cheung, D.~Tu, W.~Broniec, and M.~Riedl, ``Bringing stories
  alive: Generating interactive fiction worlds,'' \emph{Proceedings of the AAAI
  Conference on Artificial Intelligence and Interactive Digital Entertainment},
  vol.~16, no.~1, pp. 3--9, Oct. 2020.

\bibitem{riedl:saretto:aamas2003}
\BIBentryALTinterwordspacing
M.~O. Riedl, C.~Saretto, and R.~M. Young, ``Managing interaction between users
  and agents in a multi-agent storytelling environment,'' in \emph{Proceedings
  of the 2nd International Conference on Autonomous Agents and Multi-Agent
  Systems}, Melbourne, Australia, July 2003, pp. 741--748. [Online]. Available:
  \url{http://www.cc.gatech.edu/~riedl/pubs/riedl-young-aamas03.pdf}
\BIBentrySTDinterwordspacing

\bibitem{CavazzaNarrativePlanning}
M.~Cavazza, F.~Charles, and S.~J. Mead, ``Planning characters' behaviour in
  interactive storytelling,'' \emph{Comput. Animat. Virtual Worlds}, vol.~13,
  pp. 121--131, 2002.

\bibitem{FacadeSystem}
M.~Mateas and A.~Stern, ``Structuring content in the façade interactive drama
  architecture,'' in \emph{Artificial Intelligence and Interactive Digital
  Entertainment Conference}, 2005.

\bibitem{NelsonNarrativeSearch}
M.~Nelson and M.~Mateas, ``Search-based drama management in the interactive
  fiction anchorhead,'' in \emph{Artificial Intelligence and Interactive
  Digital Entertainment Conference}, 2005.

\bibitem{WeyhrauchNarrativePlanning}
P.~Weyhrauch, \emph{Guiding Interactive Drama}, ser. Research paper.\hskip 1em
  plus 0.5em minus 0.4em\relax Carnegie-Mellon University. Department of
  Computer Science, 1997.

\bibitem{autoagents}
R.~Aylett and S.~Louchart, ``If i were you: double appraisal in affective
  agents,'' in \emph{Proceedings of the 7th International Joint Conference on
  Autonomous Agents and Multiagent Systems - Volume 3}, ser. AAMAS '08.\hskip
  1em plus 0.5em minus 0.4em\relax Richland, SC: International Foundation for
  Autonomous Agents and Multiagent Systems, 2008, p. 1233–1236.

\bibitem{Scheherazade}
B.~Li and M.~O. Riedl, ``Scheherazade: Crowd-powered interactive narrative
  generation,'' in \emph{Proceedings of the 29th AAAI Conference on Artificial
  Intelligence}, 2015.

\bibitem{SwansonMLforIF}
R.~Swanson and A.~S. Gordon, ``Say anything: Using textual case-based reasoning
  to enable open-domain interactive storytelling,'' \emph{ACM Trans. Interact.
  Intell. Syst.}, vol.~2, no.~3, Sep. 2012.

\bibitem{Hartsook2011TowardSupportingStories}
\BIBentryALTinterwordspacing
K.~Hartsook, A.~Zook, S.~Das, and M.~O. Riedl, ``Toward supporting stories with
  procedurally generated game worlds,'' in \emph{Proceedings of the 2011 IEEE
  Symposium on Computational Intelligence and Games}, 2011. [Online].
  Available: \url{https://ieeexplore.ieee.org/document/6032020}
\BIBentrySTDinterwordspacing

\bibitem{Nasir2024Word2WorldGS}
M.~U. Nasir, S.~James, and J.~Togelius, ``Word2world: Generating stories and
  worlds through large language models,'' \emph{ArXiv}, vol. abs/2405.06686,
  2024.

\bibitem{Basavatia2024STARLINGST}
S.~Basavatia, K.~Murugesan, and S.~Ratnakar, ``Starling: Self-supervised
  training of text-based reinforcement learning agent with large language
  models,'' \emph{ArXiv}, vol. abs/2406.05872, 2024.

\bibitem{LLMsAndGamesSurvey}
R.~Gallotta, G.~Todd, M.~Zammit, S.~Earle, A.~Liapis, J.~Togelius, and G.~N.
  Yannakakis, ``Large language models and games: A survey and roadmap,''
  \emph{IEEE Transactions on Games}, p. 1–18, 2024.

\bibitem{TextWorld}
M.-A. C{\^o}t{\'e}, {\'A}.~K{\'a}d{\'a}r, X.~Yuan, B.~A. Kybartas, T.~Barnes,
  E.~Fine, J.~Moore, M.~J. Hausknecht, L.~E. Asri, M.~Adada, W.~Tay, and
  A.~Trischler, ``Textworld: A learning environment for text-based games,'' in
  \emph{CGW@IJCAI}, 2018.

\bibitem{Urbanek2019LearningTS}
J.~Urbanek, A.~Fan, S.~Karamcheti, S.~Jain, S.~Humeau, E.~Dinan,
  T.~Rockt{\"a}schel, D.~Kiela, A.~Szlam, and J.~Weston, ``Learning to speak
  and act in a fantasy text adventure game,'' in \emph{Conference on Empirical
  Methods in Natural Language Processing}, 2019.

\bibitem{Ye2022NeuralStoryPlanning}
\BIBentryALTinterwordspacing
A.~Ye, C.~Cui, T.~Shi, and M.~O. Riedl, ``Neural story planning,'' \emph{arXiv
  preprint arXiv:2212.08718}, 2022. [Online]. Available:
  \url{https://arxiv.org/abs/2212.08718}
\BIBentrySTDinterwordspacing

\bibitem{Kriegel2008EmergentNA}
M.~Kriegel and R.~Aylett, ``Emergent narrative as a novel framework for
  massively collaborative authoring,'' in \emph{International Conference on
  Intelligent Virtual Agents}, 2008.

\bibitem{Valevski2024DiffusionMA}
D.~Valevski, Y.~Leviathan, M.~Arar, and S.~Fruchter, ``Diffusion models are
  real-time game engines,'' \emph{ArXiv}, vol. abs/2408.14837, 2024.

\end{thebibliography}

\appendix
\section{Example of LLM Prompting for Action Parsing}
\subsection{Prompting Novel Actions}
\label{appendix:prompting-dynamic}
\noindent The following is an example of how the LLM is prompted:

\begin{quote}
You are creating a text-based adventure game similar to Zork. One of your responsibilities is to design the game engine's action system. Actions can alter the game's state represented by a tree structure with nodes. Each node can be a room, item, or character.

Given a sentence, determine the requirements of the actions, utilizing a set template. For requirements (between 1 and 3), focus on either items, attributes (like DnD), or other events that might be necessary preconditions to enable the action.

Do not include any requirements that would be considered fundamental, such as being in the same location or existing. Those are unnecessary.

Preceding events have nothing to do with items and attributes; they are independent events that must come before the input.

If the "preceding" input is true, include one preceding event.

\textbf{Possible Effects:}
\begin{enumerate}
    \item Move \texttt{\{node1\}} to \texttt{\{node2/inventory\}}
    \item Set \texttt{\{node.some\_attribute\}} to \texttt{\{value\}}
    \item Delete \texttt{\{node\}}
    \item Add \texttt{\{node\_name\}} of type \texttt{\{Item/Character\}} to \texttt{\{node/inventory\}}
    \item Display Some message with \texttt{\{node.some\_attribute\}}
\end{enumerate}
\end{quote}

\subsubsection{Example 1: Distracting the Guard}

\textbf{Input:}
\begin{quote}
Input: Distract the guard; room: at dungeon; preceding: true.
\end{quote}

\textbf{Output:}
See Fig.~\ref{fig:distract-guard}

\begin{figure}[t]
\begin{lstlisting}
{
    "input": Distract the guard,
    "output": {
        "player": "adventurer",
        "subject": "guard",
        "rooms": ["dungeon"],
        "items": [],
        "characters": ["guard"],
        "attributes": {},
        "preceding_events": ["Set bush on fire],
        "annotated_form": "{player: adventurer} distracts the {characters[0]: guard} at {rooms[0]: dungeon}.",
        "base_form": "distract the {characters[0]} at {rooms[0]}",
        "fundamental_preconditions": ["{player at rooms[0]}", "{characters[0] at rooms[0]}"],
        "additional_preconditions": [],
        "attribute_effects": ["{guard.distracted==True}"],
        "effects": ["Set {characters[0]}.distracted to True"],
        "display": "You distracted the {characters[0]}.",
        "complete_sentence": "Adventurer distracts the guard"
    }  
}
\end{lstlisting}
\caption{Example output for the dynamically generated action intermediate JSON data structure\texttt{Distract the guard}.}
\label{fig:distract-guard}
\end{figure}

\vspace{12pt}

\end{document}